# Formalizing Attack Scenario Description: A Proposed Model

Quentin GOUX

Laboratoire CEDRIC. Conservatoire National des Arts et Métiers, Paris, France – quentin.goux2.auditeur@lecnam.net

Nadira LAMMARI

Laboratoire CEDRIC. Conservatoire National des Arts et Métiers, Paris, France – lammari@cnam.fr

**Abstract**

Organizations face an ever-changing threat landscape. They must continuously dedicate significant efforts to protect their assets, making their adoption of increased cybersecurity automation inevitable. However, process automation requires formalization of input data. Through this paper, we address this need for processes that use attack scenarios as input. Among these processes, one can mention both the generation of scripts for attack simulation and training purposes, as well as the analysis of attacks. Therefore, the paper's main research contribution is a novel formal model that encompasses the attack's context description and its scenario. It is abstracted using UML class model. Once the description of our model done, we will show how it could serve an upstream attack analysis process. We will show also its use for an automatic generation of attack scripts in the context of cybersecurity training. These two uses cases constitute the second contribution of this present research work.

**Keywords:** Attack model, attack formalization, attack context formalization,

## 1 Introduction

Despite the constant evolution of cybersecurity, it remains a significant concern for stakeholders. Systematic reviews of literature quantitatively confirm the steady increase of interest in this field, whether from academics or industries. For instance, in a recent review by Ayanwale et al. (2024) focused on cybersecurity in higher education, the bibliometric indicator for trend over years clearly reveals a first increase from 2016, followed by a sharp rise since 2019. This growing trend is driven by a growing awareness of current cybersecurity challenges, including the prompt and efficient remediation of increasingly sophisticated attacks and the skill enhancement of information technology users and employees from SOCs (Security Operations Centers). As with many cybersecurity processes, remediation continuous improvement depends on the quality of the attack analysis process. It also depends on the granularity of the event sequence description that led to the attack. This description, usually named "attack scenario description", generally appears in the incident reports. It can also be used for setting up cybersecurity exercises on virtual platforms.

To build high-quality cybersecurity exercises, it is strongly recommended, in the education field, to confront trainees with realistic situations involving well-known attacks. The design and development of such exercises assume that the attack scenarios it can invoke are already specified. They require a lot of effort. They are tedious, error-prone, time and money consuming activities. Advanced skills are needed too. It would therefore be beneficial to automate these activities.

As any automation, the automation of these activities requires formalized input. However, the input required for the design of an exercise is, above all, an attack scenario. The latter, available in incident reports, is expressed at best semi-formally using attack models. Most of the time, it is conveyed in natural language. An expression of the attack scenario in natural language complicates the extraction and automatic formalization of the knowledge required for automation and does not guarantee its completeness. Furthermore, a semi-formal expression of an attack scenario through an attack model assumes that the exercise designer is proficient in the model used and its notation and that the semi-formal description contains all required knowledge, which is not always the case. Therefore, the present paper provides a formalized description of attack scenarios useful both for setting up cybersecurity exercises and for any automatic process that requires an attack description. Through formalization, we can meet automation requirements while also incorporating all knowledge that can be expressed using pre-existing attack models. Finally, to make our formalized description of the attack scenarios easier to comprehend, we use the UML class model to abstract it.

The reminder of this paper is organized as follows. Section 2 discusses related works while Section 3 describes our contribution to the formalization of attack scenarios. Section 4 provides two use cases of our formalized attack scenario model. Finally, the last section concludes the article and presents future work.

## 2 State of the art

Cyber threats are not only complex but also constantly evolving. This cyber threat landscape is forcing companies to be aware of the occurring attacks to improve their defense postures and processes. By facilitating information sharing, Cyber Threat Intelligence is greatly contributing to this awareness since it provides, among other things, various models and frameworks for describing attacks. Among them, Naik et al. (2022) isolates the three most popular ones: Lockheed Martin's Cyber Kill Chain (CKC), the MITRE ATT&CK Framework and the Diamond Model.

The CKC was initially designed to describe intrusion, Hutchins et al. (2011), then it has been generalized to any cyberattack. CKC outlines an overview of an attack's lifecycle through a fixed sequence of phases which are: "Reconnaissance", "Weaponization", "Delivery", "Exploitation", "Installation", "Command and Control" and "Actions on Objectives". Descriptions of the phases in natural language prevent the CKC from being used as they are for any automation process.

The MITRE ATT&CK Framework, Strom et al. (2020), focuses on attacker behaviors. It can be considered as a knowledge base that provides, within a matrix, a series of tactics that are broken down into techniques and then into actions that attackers can undertake to achieve their objectives. Therefore, unlike CKC, MITRE ATT&CK gives a technical understanding and description of how an attack unfolds. However, as for CKC, the description is purely informal. Thus, although more technical, MITRE ATT&CK cannot directly serve cyber process automation. A formalization effort of its valuable content is required.

The Diamond model explains that an adversary (who did the attack) exploits a capability (how did the adversary do it) over an infrastructure (what was used) against a victim (who was targeted), Naik et al. (2022). It organizes this event characterized by these fourth elements (adversary, capability, infrastructure and victim) into a diamond-shaped graph. Thus, to express an attack composed usually of a chain of causal events, Caltagirone et al. (2013) suggested to use an activity thread diagram described in their paper. This diagram is a kind of attack graph.

To serve attack modeling, academic literature provides other graph-based attack models. Lallie et al. (2020) surveys these models and highlights that the most popular ones are attack graph and attack tree. The latter is visually represented through an arborescent structure where the different ways to achieve an attack are highlighted. The first node of the tree represents the attacker's goal to be achieved. The other nodes fall into three categories: nodes representing subobjectives of parent nodes, "OR" nodes describing alternatives, and "AND" Nodes expressing steps in achieving the same subobjective.

According to Zenitani (2023), there exists 3 major kinds of attack graphs: host-based graph, state enumeration graph and exploit dependency graph. A State-enumeration graphs (SEAG) corresponds to a state-transition diagram in which the vertex represents a system state and an arc a transition between two states Dacier and Deswarte (1994); Louthan (2011); Pinchinat et al. (2014); Swiler and Phillips (1998). One of the main drawbacks of this model is the lack of a language that allows a formal description of the underlying network model associated with an SEAG representing an attack. This issue constitutes an obstacle to the utilization of this model in an automated process. A host-based attack graph (HBAG) describes a network topology where the vertex is a host labelled with their vulnerabilities, Ammann et al. (2005). Since the hosts are not previously specified, the construction of the HBAG relies on modeler's intuition for the definition of the hosts. Furthermore, this model is not supported by a controlled vocabulary for the attack description. Therefore, the modeler employs its own vocabulary which complicates its uses as an entry of any automated process. Finally, an HBAG model cannot directly represent attack progression through visualization of access graph. To overcome this disadvantage, attack modeling using HBAG model must rely on complementary information. In an exploit-dependency graph (EDAG), a vertex describes an exploit (i.e. attack) and an arc a condition. More precisely, an incoming arc is a precondition of an exploit, and the outgoing arc is its postconditions, Jajodia et al. (2005); Ruijters and Stoelinga (2015); Schneier (1999); Weiss (1991). As with other models, this model lacks formalization. Indeed, it does not provide practical ways on how to describe exploits and conditions to facilitate its exploitation by an automatic process.

To sum up, all the models proposed in the literature and those proposed by the CTI are lacking formalization, which is nowadays an essential requirement given the critical need of companies to automate their cybersecurity business processes, including those handling attack models.

## 3 Towards formalization of attack scenarios

In this section we propose our formalization of an attack scenario. We consider that this formalization to be useful must gather both the context of the attack as well as its scenario. Indeed, the attack context is constituted of all the resources involved in the attack, mainly (i) those targeted by the attack and (ii) those mobilized by the attacker to achieve his/her goal. Before the attack unfolds, its context is in a given state. Then its state evolves as the scenario progresses until it reaches a final state when the attacker goals are achieved. This progression is usually carried out in stages following actions. It is these actions that cause the context to evolve. They are undertaken by agents of which the attacker is one and he is, most often, the initiator of the scenario. The first level of abstraction of an attack scenario is depicted in Figure 1. Paragraphs 3.1 and 3.2 go deeper in describing these two artefacts of an attack, up to providing respectively their formal description.

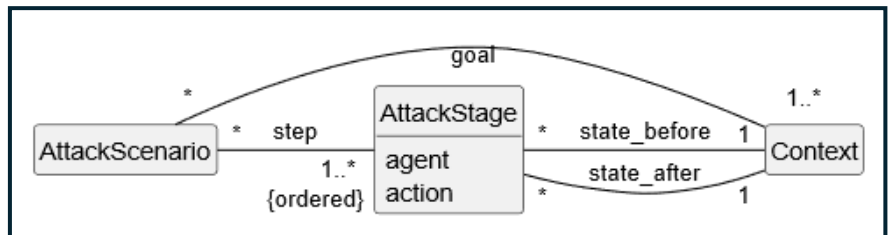
Figure 1 Attack staging conceptual model *(source: Own)*

As an illustration of our formalization of these two artefacts, we will use an example consisting of a scenario where an attacker steals the credentials that a victim uses to connect to a shopping website in order to access his/her account. In the rest of the paper, this attack will be called "SnifAttack". We will assume for this attack the following scenario steps:
- The attacker scans its local network gateway to find a listening SSH service;
- The attacker uses default credentials to take control of the router;
- The attacker causes the router to collect traffic passing through it;
- the victim sends his/her credentials to log on to the website;
- The attacker finds out the victim credentials from reading the collected traffic;
- Finally, the attacker authenticates with the victim's credentials on the shopping website.

From a cybersecurity point of view, this relies on a 4 hosts infrastructure: the attacker's system, a router, the victim's system and the shopping website (see Figure 2). Those hosts can reach one another via 3 networks: (i) the one where the attacker's system is located, (ii) the one just adjacent to the first network where the victim's system is, and (iii) the Internet for the remote website. Thus, data sent on one of those networks is forwarded by a network routing service, so it successfully reaches its destination system.

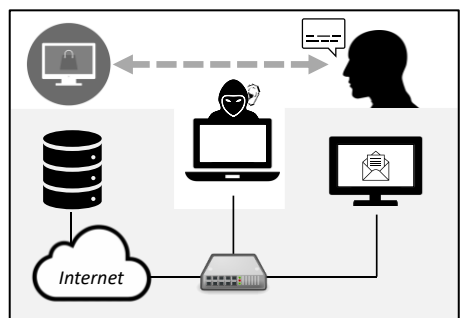
Figure 2 Graphical representation of SnifAttack *(source: Own)*

From a practical point of view, we choose to store any formal attack scenario with its context as a knowledge graph. For this purpose, we selected Neo4j as the storage system. For the sake of clarity and conciseness, we present an abstraction of the knowledge graph into a UML class diagram and occasionally give some extracts from the knowledge graph associated with the illustrative example.

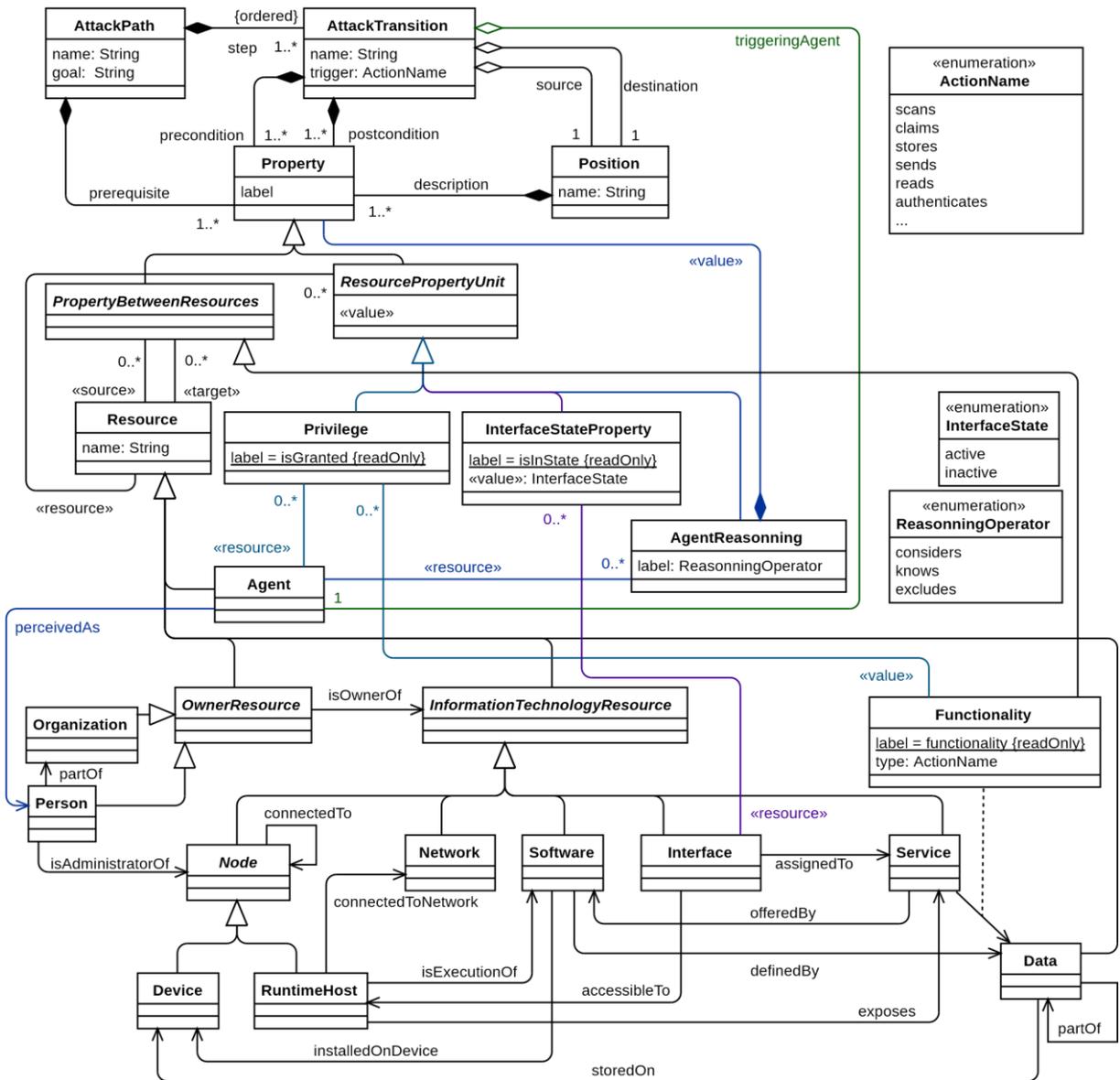

Figure 3 Our formal attack model abstracted using UML class diagram

*3.1    Attack context formalization*

Our attack context formalization is inspired, on one hand, by Wunder et al. (2011) to express cybersecurity domain assets, and, on the other hand, by the concepts of the technology layer of ArchiMate The Open Group (2022) that allows to model the relations between behaviors and structures whether they are passive or active. It is depicted in the UML conceptual model of Figure 3. It constitutes the lower part of this model.

We gather under the concept of attack context all the resources involved in the attack. Thus, we distinguish three types of assets: technological resources, their owners, the stored and/or manipulated data which could be the target of the attack, the agents participating in the attack. Attack resources are usually interconnected. To express these interconnections we use labels the majority of which are proposed in Wunder et al. (2011). They will be part of the controlled vocabulary serving to formalize the description of an attack scenario.

An agent represents any actor involved in the attack scenario, whether a conscious attacker or an actor unconsciously participating in carrying out the attack. For example, in a phishing attack, the victim who actively provides the craved sensitive information is an agent. In our illustrative example SnifAttack, the attacker and the victim that sends his/her credentials to log on to the website are both represented through the agent concept.

A Technological resource could be a network, a device, a software, a runtime-host, a service, an interface or a functionality. A device represents a physical IT resource upon which software may be installed, executed or data may be stored. We distinguish, in our model, a software from an instance of its execution that we named "runtime-host" since, during the realization of an attack, the execution of a software may produce a host that could reach remote ones through a network.

For example, in SnifAttack the router is a runtime-host since it is an execution of its software VyOS. It is connected to 3 networks including the LocalLAN in which the attacker host is also connected. Hence the router is a host reachable from the attacker one.

Devices may be plugged in one another to be physically connected. They can also be logically connected to runtime-hosts when their driver or their operating system are executed. Within software, some behavior can be implemented in order to be exposed as services via interfaces. Therefore, the interface constitutes the intended access means to an offered service. A functionality embeds an action that may be performed by a service on a data. For example, in SnifAttack the listening port is an interface assigned to the SSH service. Thus, remote runtime-hosts might attempt to establish a connection to the SSH service using the listening port. Once the attacker host is connected to the router via SSH service, it causes the router to collect traffic passing through it. Hence, the action « store » can be performed by the SSH service on the collected traffic (i.e. data).

Figure 4 is a very small excerpt of the No4j knowledge graph representing the formal specification of SnifAttack context. It expresses the fact that, in the infrastructure, the 4 hosts are connected to their networks and that the routing service is provided by the router.

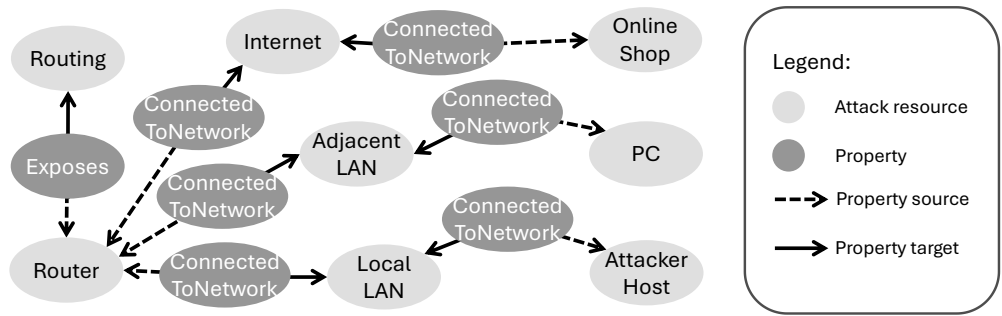

Figure 4 Excerpt of SnifAttack context specification

### 3.2 Attack scenario formalization

An attack scenario is expressed as a sequence of exploits usually called an attack path Lallie et al. (2020). According to these authors, each exploit may be defined by a composition of four elements: (1) the vulnerability that allows the exploit realization, (2) the exploit consequences, (3) the source committing the exploit and (4) the exploit target. A vulnerability is described by a set of conditions on system states, named preconditions. They express the success of the exploit. Exploit consequences are also expressed as conditions on system states, named postconditions. Some postconditions of an exploit might be preconditions of another exploit in the same attack, therefore the latter depends on the former in the same scenario. Hence, the global system state evolves with each step of the attack scenario.

Since attack graphs are the most popular attack model, our formal attack model follows their logic. In those models (specifically in the state-enumeration models), an attack path looks like a state-transition model where states correspond to the context states and where the successive evolutions of the context states are described through transitions. In other words, each change of context states following an exploit is represented by a transition.

The top part of the UML conceptual model in Figure 3 contains our proposal for the formal specification of an attack scenario. In this diagram, an attack path representing an attack scenario is named. It is associated with one or more attack objectives. Its prerequisites bring together all the required conditions for the achievement of its steps. Each step is modeled by a transition from the attack of a source position to a destination position, for which a position fully describes a state of its context. The transition is associated with not only preconditions that provide information on the state of the context necessary upstream of the realization of the transition but also postconditions that are satisfied once the transition is realized. For instance, in SnifAttack, we must express all the prerequisites. Among them, we have to express FACT1 of Figure 5A. Likewise, we have to express all the preconditions of each transition. Among preconditions of the transition corresponding to SnifAttack's first step, we will find the one describing FACT2 of Figure 5A. Among preconditions of the transition corresponding to SnifAttack's fifth step, we will find the property expressing FACT3 of Figure 5A.

The trigger corresponds to the action that triggers the transition. This action is performed by an agent that can be either the attacker or any actor, even one from the targeted. The names associated with the actions belong to an enumeration.

Given that a system is above all constituted of all the resources constituting the attack infrastructure, Li et al. (2022) suggested to express postconditions and preconditions using two kind of resources properties : those characterizing resources and those describing the logical or physical connection between them. For the description of prerequisites of an attach path, the pre and postconditions of its transitions and its positions, we opted for this suggestion while expressing the properties in our own formalism. Thus, property describing relationships between attack resources is named in Figure

3 "Property Between Resources". It has a source and a target resource. Its label is part of the controlled vocabulary expressed in Figure 3. As an example, the instantiation of our conceptual model for FACT1 is depicted in Figure 5B.

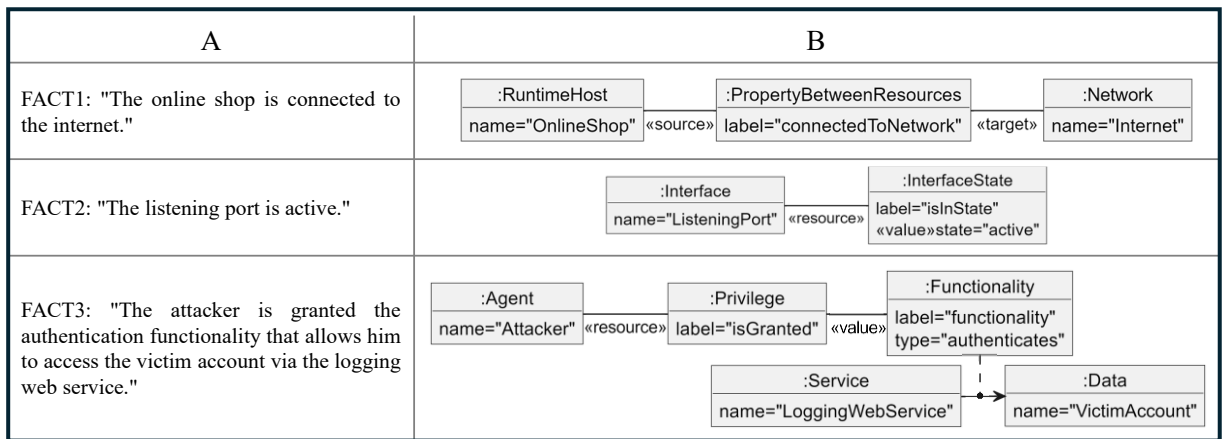

Figure 5 Examples of SnifAttack context fact

A Property characterizing a resource is named, "Resource Property Unit". Characterization could be a value. For an interface type property, named in our model "Interface State Property", the only label that could be assigned to it is "isInState" and its value could be either "inactive" or "active". As an example, the instantiation of our conceptual model for FACT2 is depicted in Figure 5B. The corresponding fragment of SnifAttack's knowledge graph is provided in Figure 7A.

Through this kind of property, we can also express privileges granted to agents. In this case, the label of this property is "IsGranted" and the characterization (value) is the granted functionality. In the case of SnifAttack, we can highlight the functionality of authentication required for accessing the victim account. This functionality is granted to the attacker in a late stage of the attack. The instantiation of our conceptual model for this fact (FACT3), is depicted in Figure 5B. The corresponding fragment of SnifAttack's knowledge graph is provides in Figure 7B.

Finally, the "Resource Property Unit", enables expressing the attacker reasoning, which includes the assumptions made prior to and during the attack, as well as the conclusions reached during the attack steps. In this case, the property is labeled with an operator from the "Agent Reasoning Operator" enumeration, and its value is a fact about the context. For example, in SnifAttack, the attacker scans its local network gateway which allows him/her to discover a listening port. Then, the attacker assumes that the listening port corresponds to the SSH service and that the SSH service is exposed by the router. The first conclusion is formulated in Figure 6A into FACT4. The instantiation of our conceptual model for this fact is also depicted in Figure 6B. Figure 7C gives the corresponding portion of SnifAttack's knowledge graph.

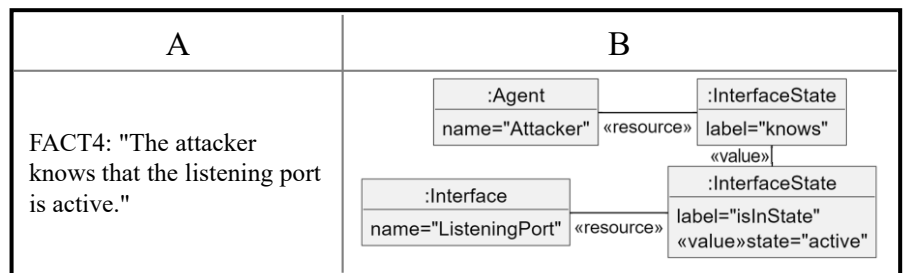

Figure 6 An example of a resource property expressing the attacker reasoning

Let us mention that even if these attacker assumptions and conclusions do not serve all automated processes, it may help explain changes in context state. They also might even be the cause of the attack key steps. Thus, they are useful for the construction of narratives. Therefore, to facilitate cybersecurity experts the construction of their narratives for the specification of an attack scenario, we chose to represent this kind of properties in our formal model.

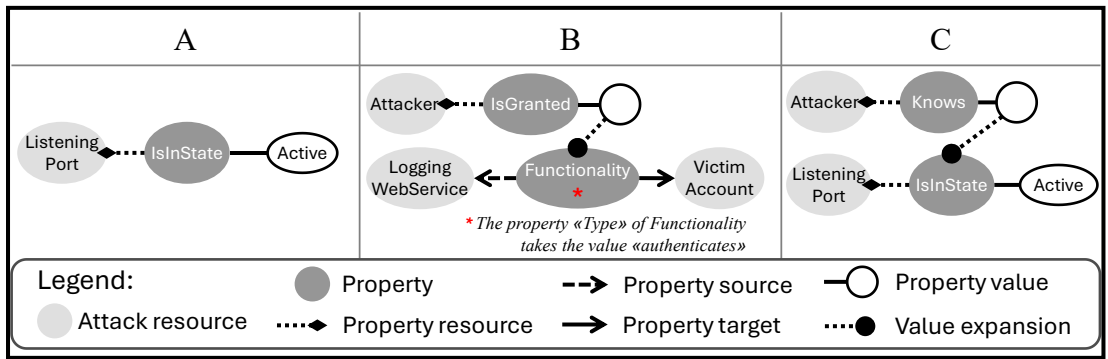

Figure 7 Portions of SnifAttack's knowledge graph

## 4 Potential usage of our formalized specification of attack scenarios

The literature is abundant with contributions that enable both the generation of attack graphs (such as the survey of Konsta et al. (2024)) and their analysis (such as Zeng et al. (2019)). The automatic analysis of attack graph content is a valuable tool for organizations. However, due to the lack of formalization of these well-known models, these analyses require beforehand the implementation of complex Natural Language Processing techniques. Since the techniques used don't consider semantic and syntactic heterogeneity of attack models, analysis processes are less generic (that is to say they are specific to a kind of attack model). We believe that running automatic analysis on a formal model such as our model will increase the genericity of these processes. To achieve this task, all we have to do is design model transformation processes that takes input from an existing attack model and produces, using transformation rules, a formalized attack model as the output. Through paragraph 4.2 of this section, we show a use case of our formal model to serve an upstream analysis process. Paragraph 4.1 is dedicated to another usage of our formal model which is the automatic generation of attack script.

### 4.1 Attack script generation

The setting up of most cybersecurity training exercises requires the availability of attack scripts. The latter, if not available, are usually produced manually from attack descriptions. By automating their production, effort and errors can be reduced. Thus, as part of our research, we proposed a process to automatically generate an attack script from a scenario and its context expressed both formally according to the model described in 3.1. Our Process is a model driven process founded on MDA principles.

As a reminder, MDA, as a variant of MDE (Model driven engineering), associates to each level a model called respectively a computation-independent model (CIM), a platform-independent model (PIM) and a platform-specific model (PSM). It also recommends the implementation of transformation mechanisms allowing the transition from a source model to a target model.

Our formal model described in Section 3 and depicted in Figure 3 constitutes the CIM model. Therefore, the first step of our process consists of the instantiation of this CIM model. The second step produces, in compliance with the instance of CIM, an attack script abstraction, named abstract script, and an attack context abstraction, named abstract infrastructure. These two artefacts constitute the instance of the PIM model, from which different PSMs can be derived. They are abstract since their description is devoid of elements relating to their implementation.

Our PIM model is expressed using the standard OASIS TOSCA (2013). Rules transforming a CIM instance into a PIM one have been designed. The PSM corresponds to an executable script whose instructions can be interpreted by the dedicated execution environment and to a concrete infrastructure on which the executable script will be applied.

The process we designed has been implemented. For the execution of attack scripts, we built our own dedicated platform mainly combining two useful tools. The first one is the OpenTOSCA ecosystem, Zimmermann et al. (2018), which initiates the provisioning of concrete infrastructures from a template topology. The second one is the automation engine Ansible, Freeman et al. (2023), which is mainly charged with the execution of the attack script.

```
topology_template:
    node_templates:
        Router: …
        AttackerHost:
            type: HostSystem
            requirements:
                - local_storage:
                    node: AttackerDevice
                …
        PortScanner:
            type: tosca.nodes.SoftwareComponent
            requirements:
                - host: AttackerHost
        LocalLAN:
            type: tosca.nodes.network.Network
            …
        AttackerHost_connectedToNetwork_LocalLAN:
            type: tosca.nodes.network.Port
            requirements:
                - link: LocalLAN
                - binding: AttackerHost
        Router_connectedToNetwork_LocalLAN: …
        …
    workflows:
        AbstractScript:
            description: 'An attacker steals the credentials that a victim uses to connect to a shopping web
            steps:
                Scan:
                    activities:
                        - call_operation: action.scans
                    on_success: [ UseOfDefaults ]
                    target: AttackerHost
                UseOfDefaults: …
                Sniffing: …
                Disclosure: …
                Discovery: …
                Checkmate: …
```

Figure 8 Excerpt of YAML file encoding the PIM of SnifAttack in TOSCA

As an example, we specified SnifAttack scenario described in Section 3. From the specification of SnifAttack stored as a Neo4j knowledge graph, the TOSCA infrastructure topology representing the abstract infrastructure is automatically generated. It comprises 18 node templates. The TOSCA workflow describing the abstract script is also automatically generated. The two artefacts are encoded in a YAML file. An excerpt of SnifAttack YAML file encoding the PIM in TOSCA is depicted in Figure 8. Finally, the infrastructure was physically instantiated, and the script automatically executed. Figure 9 supplies the command line produced from an automated execution of SnifAttack.

```
$ ansible-playbook -i attackPSM_env/AAE-OpenTOSCA_inst-2614/00_inventory.yaml attackPSM_playlib/AttackScript.yaml

PLAY [Scan (Attacker scans) - The attacker scans its local network gateway, then finds a listening SSH service.] ***

TASK [AttackTransition_Scan : scans] ******************************************
ok: [AttackerHost]

PLAY [UseOfDefaults (Attacker claims) - The attacker uses default cedentials to take control of the router.] ***

TASK [AttackTransition_UseOfDefaults : claims] ********************************
ok: [AttackerHost]

PLAY [Sniffing (Attacker stores) - The attacker has the router do the collecting of all traffic passing through.] ***

TASK [AttackTransition_Sniffing : stores] *************************************
changed: [AttackerHost]

PLAY [Disclosure (ActingVictim sends) - The victim sends his/her credentials to log on the website.] ***

TASK [AttackTransition_Disclosure : sends] ************************************
changed: [PC]

PLAY [Discovery (Attacker reads) - The attacker finds out the victim`s credentials from reading the collected traffic.] ***

TASK [AttackTransition_Discovery : --- internal: retrieves a local copy of the collected trafic's dump file ---] ***
changed: [AttackerHost]

TASK [AttackTransition_Discovery : reads] *************************************
ok: [AttackerHost]

PLAY [Checkmate (Attacker authenticates) - The attacker authenticates with the victim`s credentials on the website.] ***

TASK [AttackTransition_Checkmate : authenticates] *****************************
changed: [AttackerHost]

PLAY RECAP ********************************************************************
AttackerHost               : ok=6    changed=3    unreachable=0    failed=0    skipped=0    rescued=0    ignored=0
PC                         : ok=1    changed=1    unreachable=0    failed=0    skipped=0    rescued=0    ignored=0
```

Figure 9 Command line output from an automated execution of SnifAttack

*4.2    A model transformation use case*

To allow analysis processes of attack scenarios to be less dependent on the existing attack models, we have suggested in the introduction of our section the implementation of model transformation mechanisms that can be used upstream of these analysis processes. Thus, an analysis process using formalized attack scenarios will benefit from a larger dataset, especially if its proper functioning and the quality of its analysis results depend on it.

There may be several types of model transformation mechanisms: those that allow us to transform a scenario expressed according to an existing model into a scenario that matches our formalization and those which act in the opposite way. In this paragraph we focus on the first ones by presenting the transformation that translates an attack path described using the formalism of Pinchinat et al. (2014). For the sake of brevity, Figure 10 shows the transformation rules of a visual representation of an attack path from the model of Pinchinat et al. (2014) into a knowledge graph corresponding to our formalized model. The first two columns give respectively the elements of the graphical syntax used in Pinchinat et al. (2014) to express an attack path and their meaning. The last column supplies its corresponding expression in Neo4j knowledge graph. As an example, Figure 11 shows the representation of SnifAttack using the notation of Pinchinat et al. (2014). The application of the transformation rules described in Figure 10 leads to the Neo4j knowledge graph of Figure 12.

Let us mention that some models like this of Pinchinat et al. (2014) do not allow a precise description of the attack. Therefore, once the transformation is completed, the generated knowledge graph needs to be enriched with useful information for analysis.

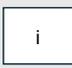

Figure 10 Transformation rules from attack model of Pinchinat et al. (2014) to our formalized description

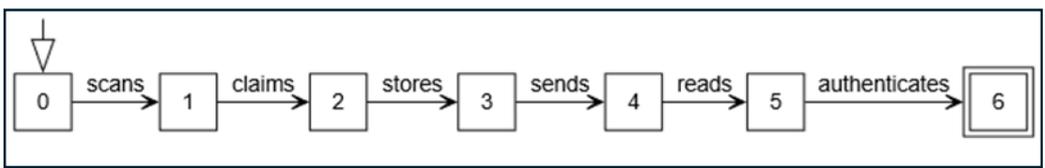

Figure 11 Visual representation of SnifAttack according to the notation of Pinchinat et al. (2014)

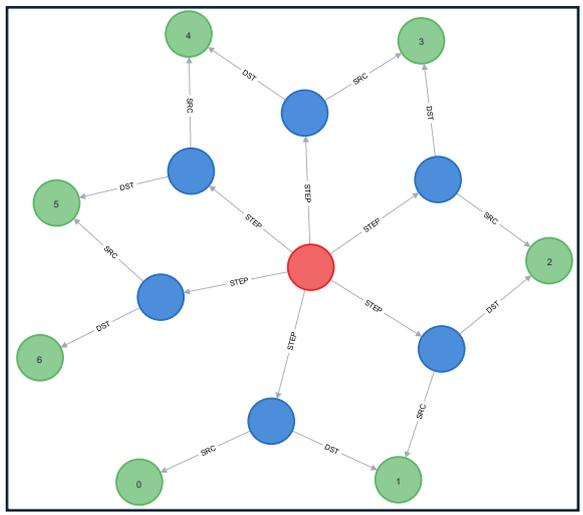

Figure 12 The Neo4j Knowledge graph corresponding to the attack scenario of Figure 11

## 5  Conclusion

The research work presented in this paper addresses the problem of formalizing cyber-attacks to ease their exploitation by automated processes. More precisely, we have been particularly interested in formalizing the description of both attack contexts and their scenarios. Thus, we proposed a detailed description of these two artifacts. We then discussed two use cases of our contribution in the context of cybersecurity automation.

Currently, we are working on the design and implementation of a user-friendly GUI to assist security experts in specifying an attack scenario according to the formalism we proposed in this article. We subsequently plan to conceive and implement automated model transformation mechanisms to move from semi-formal attack scenario specifications to formal ones. Further works will concern automatic construction of attack graphs by reuse of existing ones.

# References

Ammann, P., Pamula, J., Ritchey, R., Street, J., (2005), 'A host-based approach to network attack chaining analysis,' Proceedings of the 21st Annual Computer Security Applications Conference (ACSAC'05), https://doi.org/10.1109/CSAC.2005.6

Ayanwale, M.A., Adeoba, M.I., Adelana, O.P., Lawal, R.O., Makhetha, I.M., Mochekele, M., (2024), 'Cybersecurity for Educational Excellence: Bibliometric Insights from Higher Education,' Proceedings of the 5th International Conference on Electro-Computing Technologies for Humanity (NIGERCON), *IEEE*, 1–8, https://doi.org/10.1109/NIGERCON62786.2024.10927315

Caltagirone, S., Pendergast, A., Betz, C., (2013), 'The Diamond Model of Intrusion Analysis (Technical Report),'

Dacier, M., Deswarte, Y., (1994), 'Privilege graph: An extension to the typed access matrix model,' Gollmann, D. (Ed.), Proceedings of the Third European Symposium on Research in Computer Security (ESORICS 94), *Springer Berlin Heidelberg*, 319–334, https://doi.org/10.1007/3-540-58618-0_72

Freeman, J., Locati, F.A., Oh, D., (2023), 'Practical ansible: learn how to automate infrastructure, manage configuration, and deploy applications, 2nd ed. ed,' *Packt Publishing, Place of publication not identified*

Hutchins, E.M., Cloppert, M.J., Amin, R.M., (2011), 'Intelligence-Driven Computer Network Defense Informed by Analysis of Adversary Campaigns and Intrusion Kill Chains,' Proceedings of the 6th International Conference on Information Warfare and Security, Leading Issues in Information Warfare & Security (ICIW), 113–125

Jajodia, S., Noel, S., O'Berry, B., (2005), 'Topological Analysis of Network Attack Vulnerability,' Kumar, V., Srivastava, J., Lazarevic, A. (Eds.), Proceedings of the Managing Cyber Threats, Massive Computing *Springer, Boston, MA, New York*, 247–266

Konsta, A.-M., Lluch Lafuente, A., Spiga, B., Dragoni, N., (2024), 'Survey: Automatic generation of attack trees and attack graphs,' *Comput. Secur.*, 137, https://doi.org/10.1016/j.cose.2023.103602

Lallie, H.S., Debattista, K., Bal, J., (2020), 'A review of attack graph and attack tree visual syntax in cyber security,' *Comput. Sci. Rev.*, 35, https://doi.org/10.1016/j.cosrev.2019.100219

Li, M., Hawrylak, P., Hale, J., (2022), 'Strategies for Practical Hybrid Attack Graph Generation and Analysis,' *Digit. Threats Res. Pract.*, 3, 1–24, https://doi.org/10.1145/3491257

Louthan, G.R., (2011), 'HYBRID ATTACK GRAPHS FOR MODELING CYBER-PHYSICAL SYSTEMS (Master of Science in the Discipline of Computer Science),' *University of Tulsa*

Naik, N., Jenkins, P., Grace, P., Song, J., (2022), 'Comparing Attack Models for IT Systems: Lockheed Martin's Cyber Kill Chain, MITRE ATT&CK Framework and Diamond Model,' Proceedings of the International Symposium on Systems Engineering (ISSE), *IEEE*, 1–7, https://doi.org/10.1109/ISSE54508.2022.10005490

OASIS TOSCA, (2013), 'Topology and Orchestration Specification for Cloud Applications Version 1.0,' *OASIS Standard*, http://docs.oasis-open.org/tosca/TOSCA/v1.0/os/TOSCA-v1.0-os.html

Pinchinat, S., Acher, M., Vojtisek, D., (2014), 'Towards Synthesis of Attack Trees for Supporting Computer-Aided Risk Analysis,' Canal, C., Idani, A. (Eds.), Software Engineering and Formal Methods (SEFM 2014) LNPSE *Springer, Cham*, 363–375, https://doi.org/10.1007/978-3-319-15201-1_24

Ruijters, E., Stoelinga, M., (2015), 'Fault tree analysis,' *Comput. Sci. Rev.*, 15–16, 29–62, https://doi.org/10.1016/j.cosrev.2015.03.001

Schneier, B., (1999), 'Attack trees,' *Dr Dobb's J.*, 24, 21–29

Strom, B.E., Applebaum, A., Miller, D.P., Nickels, K.C., Pennington, A.G., Thomas, C.B., (2020), 'MITRE ATT&CK®: Design and Philosophy,'

Swiler, L.P., Phillips, C., (1998), 'A graph-based system for network-vulnerability analysis,' *Sandia National Lab. (SNL-NM), Albuquerque, NM (United States)*, https://doi.org/10.2172/573291

The Open Group, (2022), 'ArchiMate® 3.2 Specification,' *Open Group*, https://publications.opengroup.org/standards/archimate/specifications/c226

Weiss, J.D., (1991), 'A system security engineering process,' Proceedings of the 14th National Computer Security Conference 572–581

Wunder, J., Halbardier, A., Waltermire, D., (2011), 'Specification for asset identification 1.1,' *National Institute of Standards and Technology, Gaithersburg, MD*, https://doi.org/10.6028/NIST.IR.7693

Zeng, J., Wu, S., Chen, Y., Zeng, R., Wu, C., (2019), 'Survey of Attack Graph Analysis Methods from the Perspective of Data and Knowledge Processing,' *Secur. Commun. Netw.*, 2019, 1–16, https://doi.org/10.1155/2019/2031063

Zenitani, K., (2023), 'Attack graph analysis: An explanatory guide,' *Comput. Secur.*, 126, 103081, https://doi.org/10.1016/j.cose.2022.103081

Zimmermann, M., Breitenbücher, U., Leymann, F., (2018), 'A Method and Programming Model for Developing Interacting Cloud Applications Based on the TOSCA Standard,' Hammoudi, S., Śmiałek, M., Camp, O., Filipe, J. (Eds.), Proceedings of the Enterprise Information Systems, LNBIP *Springer, Cham*, 265–290, https://doi.org/10.1007/978-3-319-93375-7_13